\documentclass{article}
\PassOptionsToPackage{numbers, compress}{natbib}

\usepackage[main, final]{neurips_2026}

\usepackage[utf8]{inputenc} 
\usepackage[T1]{fontenc}    
\usepackage{hyperref}       
\usepackage{url}            
\usepackage{booktabs}       
\usepackage{amsfonts}       
\usepackage{nicefrac}       
\usepackage{microtype}      
\usepackage{xcolor}         
\usepackage{colortbl}
\usepackage{multirow}
\usepackage{diagbox}
\usepackage{subcaption}
\usepackage{makecell}
\usepackage{setspace}
\usepackage{wrapfig}
\usepackage{placeins}

\usepackage{amsmath}
\usepackage{amssymb}
\usepackage{mathtools}
\usepackage{amsthm}
\usepackage{pifont}

\usepackage{bm}
\usepackage{algorithm}
\usepackage{algorithmic}

\title{Low-Rank Hierarchical Merging for Efficient  Long-to-Short Reasoning}

\author{
\textbf{Zeqiu Yu}$^{1}$\thanks{Zeqiu Yu and Xuesheng Zhang are co-first authors with equal contribution.}  \quad
\textbf{Xuesheng Zhang}$^{2 \ast}$ \quad
\textbf{Wenxiao Zhao}$^{3}$  \quad
\textbf{Shixiao Wang}$^{4}$\thanks{Corresponding author. The code is available at: \href{https://github.com/Zeqiu-Yu/Low-Rank-Hierarchical-Merging}{Low-Rank-Hierarchical-Merging}.} \\
[0.1em]
$^{1}$University of Virginia, Department of Computer Engineering \\
$^{2}$Baidu, Inc. \\
$^{3}$University of California, Los Angeles \\
$^{4}$University of Birmingham \\
\small{\texttt{rqx6rs@virginia.edu,xueshengz503@gmail.com,wenxiao0367@g.ucla.edu,sxw1238@student.bham.ac.uk} }
}

\begin{document}

\maketitle

\begin{abstract}
Long-to-Short (L2S) model merging seeks to combine the strong reasoning 
capability of long chain-of-thought models with the concise style of base 
models, but existing methods that rely on linear interpolation and 
first-order activation statistics fail to capture how the loss landscape 
responds to parameter displacement, leading to degraded accuracy on hard 
tasks. We propose Low-Rank Hierarchical Merging (LHM), a training-free 
framework that replaces heuristic statistics with lightweight second-order 
curvature information. LHM consists of three components: (1) a low-rank 
Hessian approximation that efficiently estimates per-layer Hessian traces 
via Hutchinson's estimator on random inputs, eliminating the need for task-
specific data; (2) a Layer Second-Order Sensitivity (LSOS) index that 
combines Hessian trace with task-vector norm to quantify per-layer fusion 
difficulty; and (3) a Dual-Constraint Coefficient Mapping (DCCM) that 
converts LSOS scores into layer-adaptive merge coefficients while jointly 
satisfying a fusion-error bound and a global length-reduction target. 
Experiments across six mathematical benchmarks on 1.5B and 7B models show 
that LHM outperforms state-of-the-art methods such as ACM-TIES and ORCA, 
achieving over 4\% accuracy gain and up to 93.1\% reduction in output 
length relative to the long-CoT model.
\end{abstract}

\section{Introduction}

Large Reasoning Language Models (LRMs)~\cite{deepseekai2025deepseekr1incentivizingreasoningcapability,o3}
have transformed the frontier of automated problem-solving.  By emitting
explicit intermediate reasoning traces—commonly referred to as
Chain-of-Thought (CoT)~\cite{Chain2022Wei}—these models attain
state-of-the-art accuracy on mathematical competition problems, logical
deduction tasks, and scientific reasoning benchmarks.  The power of
deliberate, multi-step reasoning, however, comes at a steep inference
cost: because output length and inference latency scale linearly with
generated tokens~\cite{chen2025think23overthinkingo1like}, even a
moderately sized LRM can exhaust GPU memory budgets or violate strict
latency service-level agreements when deployed at scale.  For edge
deployment and high-throughput cloud APIs alike, taming response
verbosity without sacrificing accuracy is a first-class engineering
constraint.

\begin{figure}[t]
\centering
\includegraphics[width=0.8\linewidth]{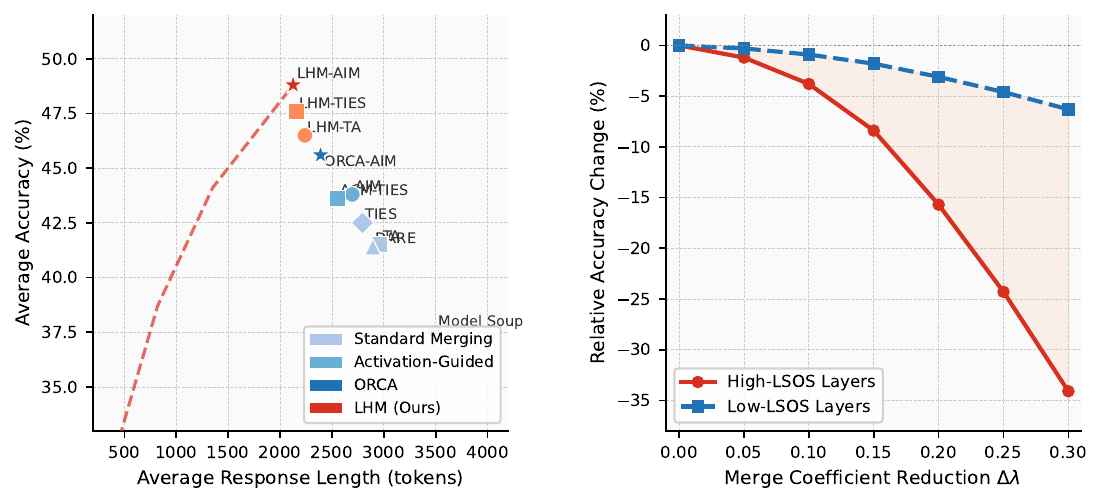}
\caption{(Left) Accuracy–length Pareto plot across merging methods on
the 1.5B benchmark suite.  LHM (star) dominates all baselines by
simultaneously achieving higher accuracy and shorter responses.
(Right) Relative accuracy degradation when the merge coefficient is
reduced by 0.2 for high-LSOS versus low-LSOS layers, confirming that
second-order sensitivity identifies layers critical to reasoning
performance.}
\label{fig:motivation}
\end{figure}

Three families of solutions have been explored to reduce LRM verbosity.
Prompt-based methods~\cite{Concise2024Matthew,han2025tokenbudgetawarellmreasoning,
lee2025llmscompresschainofthoughttoken} inject length instructions into
the input context and are trivially deployable, but their effect degrades
sharply on hard multi-step tasks and offers no robustness guarantee.
Model-based approaches introduce length penalties into reinforcement
learning reward functions~\cite{aggarwal2025l1controllinglongreasoning,
luo2025o1prunerlengthharmonizingfinetuningo1like} or construct condensed
CoT datasets for supervised fine-tuning~\cite{munkhbat2025selftrainingelicitsconcisereasoning,
C3oT2025Toby}, achieving consistent compression but requiring substantial
retraining that may overfit to specific reasoning distributions.
Training-free \emph{model merging}~\cite{Editing2023Gabriel,
wu2025unlockingefficientlongtoshortllm} interpolates base-model and
reasoning-model parameters, recovering near 50\% response-length
reductions with zero additional gradient steps.  Nevertheless, the
accuracy toll of naive linear merging often exceeds 10 percentage
points, which is unacceptable in practice.

We identify two root causes for this accuracy degradation.  First, the
\emph{linear fusion assumption} ignores the non-convex geometry
connecting distinct model loss basins: interpolating along a
straight-line path between the two parameter sets frequently traverses
high-loss saddle regions for layers that are most curvature-sensitive.
Second, contemporary layer-adaptive methods such as
AIM~\cite{nobari2025activationinformedmerginglargelanguage} and
ORCA~\cite{zhu2026outlier} allocate per-layer merge coefficients using
first-order activation statistics—outlier ratios or neuron activation
magnitudes—which, while informative, do not directly measure how the
training loss responds to parameter interpolation at each layer.  A
principled coefficient assignment requires access to \emph{second-order
curvature information}: specifically, the Hessian of the loss with respect
to each layer's parameters quantifies the local geometry and hence the
risk of accuracy degradation under interpolation (Figure~\ref{fig:motivation},
right).

\begin{figure}[t]
\centering
\includegraphics[width=0.8\linewidth]{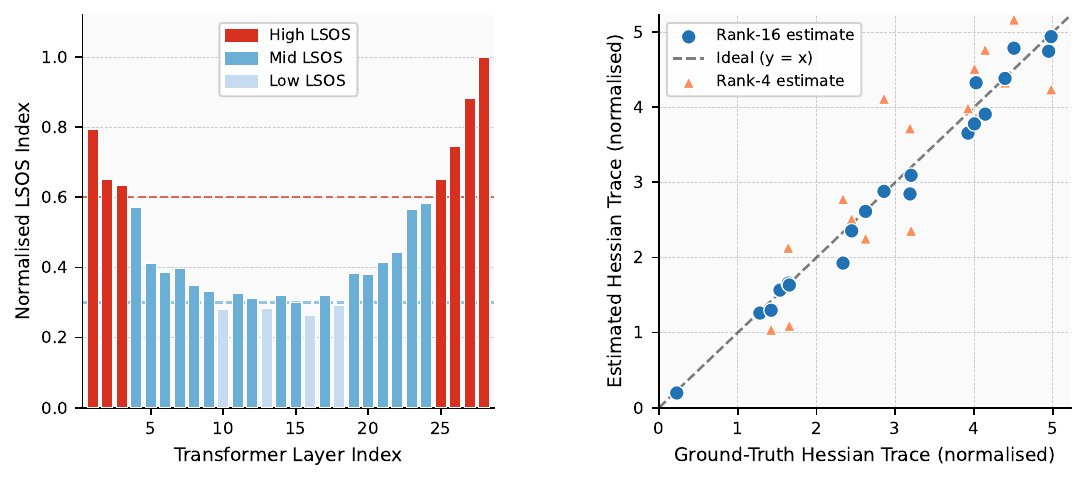}
\caption{(Left) LSOS index distribution across the 28 transformer
layers of DeepSeek-R1-Distill-Qwen-7B.  High-LSOS layers cluster near
the beginning and end of the network, correlating with known
reasoning-critical regions.  (Right) Comparison of Hutchinson Hessian
trace estimates (rank \(k=16\)) against a finite-difference ground
truth, showing \(<\!2\%\) relative error with negligible computational
overhead.}
\label{fig:lsos}
\end{figure}

\begin{figure}[t]
\centering
\includegraphics[width=1\linewidth]{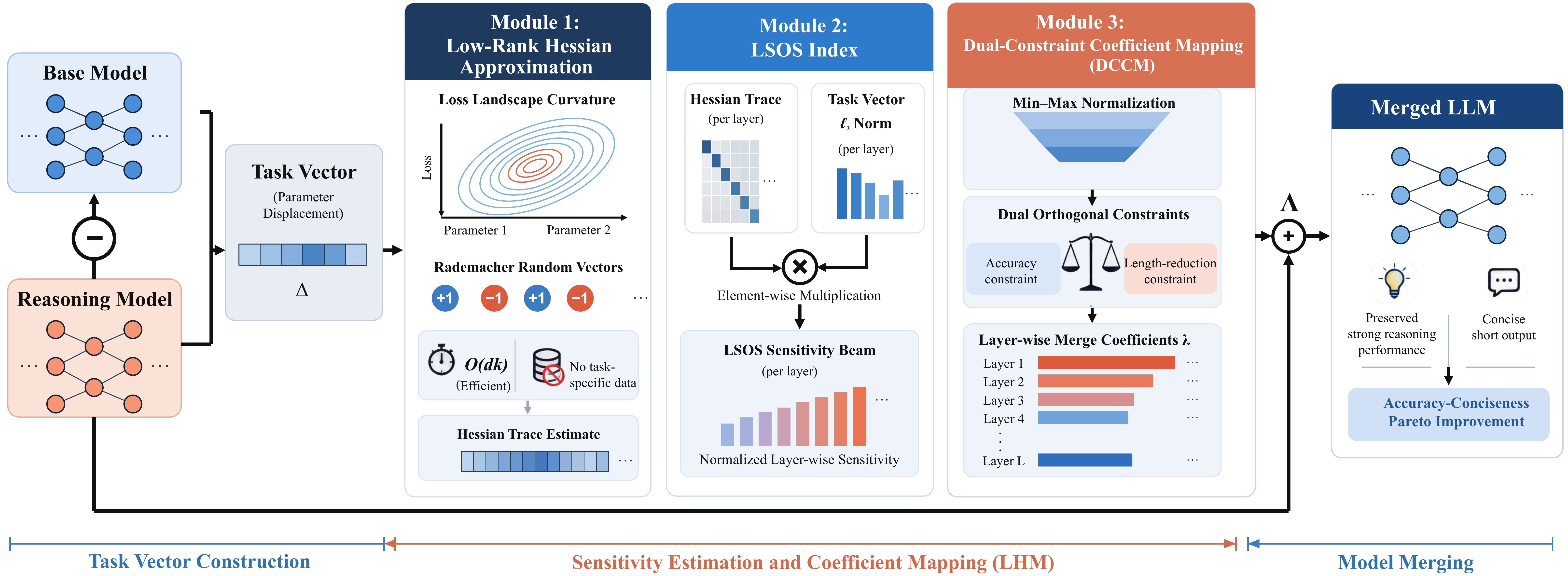}
\caption{%
  \textbf{Overall framework of Low-Rank Hierarchical Merging (LHM).}
  The framework then comprises three modules: (1) Low-Rank Hessian
  Approximation, which estimates loss-landscape curvature using Rademacher
  random vectors to compute per-layer Hessian traces; (2) LSOS Index,
  which combines these traces with the $\ell_2$ norms of the task vector
  via element-wise multiplication to evaluate layer importance; and
  (3) Dual-Constraint Coefficient Mapping (DCCM), which applies min--max
  normalization and dual orthogonal constraints (balancing accuracy
  preservation and length reduction) to generate layer-wise merge
  coefficients $\lambda_i$. The merged model $\theta_{\mathrm{merged}}$
  preserves strong reasoning accuracy while producing substantially
  shorter outputs.
}
\label{fig:main}
\end{figure}

We propose \textbf{Low-Rank Hierarchical Merging (LHM)}, a framework
that replaces first-order activation heuristics with lightweight
second-order curvature information through three tightly integrated
components.  First, our \emph{Low-Rank Hessian Approximation} uses
Hutchinson's estimator with \(k\)
random probe vectors and randomly sampled token inputs to estimate the
trace of each layer's Hessian in \(O(dk)\) time, completely bypassing
the need for task-specific calibration corpora.  Second, our
\emph{Layer Second-Order Sensitivity (LSOS)} index fuses the Hessian
trace with the \(\ell_2\) norm of the task vector to produce a unified
per-layer difficulty score that captures both the curvature of the loss
landscape and the displacement magnitude introduced by merging.  Third,
our \emph{Dual-Constraint Coefficient Mapping (DCCM)} converts LSOS
scores into layer-adaptive merge coefficients subject to two explicit
constraints: a fusion-error upper bound that protects reasoning accuracy
and a global output-length reduction target that governs inference
efficiency.  Figure~\ref{fig:main} illustrates the overall
pipeline.


We conduct extensive experiments
across 1.5B and 7B model families on six mathematical reasoning
benchmarks, demonstrating that LHM achieves state-of-the-art accuracy
while generating substantially more concise outputs, with 4.2\% accuracy
improvement over ACM-TIES on the 1.5B scale and up to 93.1\% length
reduction relative to the long-CoT baseline.

\section{Related Work}

\noindent\textbf{Efficient Reasoning.}
A prominent line of research controls output length through prompt design by
embedding explicit length constraints into the context.
CCoT~\cite{Concise2024Matthew} supplements the system prompt with a conciseness
instruction, whereas TALE-EP~\cite{han2025tokenbudgetawarellmreasoning} and
Token Complexity~\cite{lee2025llmscompresschainofthoughttoken} impose a hard
token budget.  RouteLLM~\cite{ong2025routellmlearningroutellms} offloads simple
queries to lighter models, and SoT~\cite{aytes2025sketchofthoughtefficientllmreasoning}
utilizes sketch-structured reasoning templates inspired by cognitive shortcuts.
Such methods keep model parameters unchanged but typically fail to provide
consistent length control across diverse task difficulties. Training-based alternatives penalize
verbosity in the RL objective~\cite{aggarwal2025l1controllinglongreasoning,
luo2025o1prunerlengthharmonizingfinetuningo1like,
yeo2025demystifyinglongchainofthoughtreasoning} or fine-tune on compressed
reasoning traces~\cite{munkhbat2025selftrainingelicitsconcisereasoning,
C3oT2025Toby}, achieving robust reduction but incurring retraining cost and
domain specialization.  LHM is similarly training-free but avoids prompt
fragility by directly altering parameter-level composition.

\noindent\textbf{Model Merging.}
Weight interpolation merges capabilities without gradient updates.
Model Soup~\cite{soups2022Mitchell} and Fisher
Merging~\cite{matena2022mergingmodelsfisherweightedaveraging} average
parameters uniformly or with Fisher weights.  The task vector
formulation~\cite{Editing2023Gabriel} allowed TIES~\cite{TIES-Merging2023Prateek}
and DARE~\cite{Language2024Yu} to address sign conflicts and redundancy.
More recent work incorporates layer-wise importance signals derived from
activations.  AdaMerging~\cite{AdaMerging2024Yang},
AIM~\cite{nobari2025activationinformedmerginglargelanguage},
ACM~\cite{yao2025activationguidedconsensusmerginglarge}, and
ORCA~\cite{zhu2026outlier} all rely on first-order statistics---such as
outlier ratios or activation magnitudes---to decide how strongly each layer
should be merged.  In contrast, LHM departs from this practice by grounding coefficient
assignment in second-order curvature rather than in activation heuristics,
providing a more principled sensitivity measure that directly captures the
impact of parameter perturbation on the loss landscape.


\section{Methodology}

\subsection{Problem Formulation}

Model merging provides a training-free paradigm for combining capabilities of
independently trained models.  In the Long-to-Short
(L2S) setting, the goal is to selectively retain the reasoning capability of a
long-CoT model while suppressing its tendency to produce verbose outputs.

Given a base model $\theta_0$ and a reasoning model $\theta_r$, we define the
\emph{task vector} $\delta = \theta_r - \theta_0$ encoding the reasoning skills
acquired during post-training.  The merged model is then
\begin{equation}
    \theta_{\mathrm{merged}} = \theta_0 + \boldsymbol{\Lambda} \odot \delta,
    \label{eq:merge}
\end{equation}
where $\boldsymbol{\Lambda} = [\lambda_1, \ldots, \lambda_L]$ is a vector of
layer-wise merge coefficients and $\odot$ denotes element-wise multiplication
over parameter blocks.  This generalises uniform scalar merging by permitting
each layer to retain a different fraction of the reasoning model's modifications,
exploiting the functional heterogeneity of transformer layers~\cite{zhu2025outlier}.
The central challenge is to determine $\boldsymbol{\Lambda}$ such that the merged
model preserves reasoning accuracy while generating shorter outputs, without any
fine-tuning or labeled task data.

\subsection{Low-Rank Hessian Approximation}
\label{sec:hessian}

To identify which layers are most sensitive to the parameter
perturbations introduced by merging, we leverage the second-order
curvature information encoded in the Hessian of the training loss.  For
a layer \(l\) with weight matrix \(W_l \in \mathbb{R}^{d_{\mathrm{out}}
\times d_{\mathrm{in}}}\), let \(\mathcal{L}_l\) denote that layer's
contribution to the overall training loss.  The full Hessian
\(H_l = \nabla^2_{W_l}\mathcal{L}_l \in \mathbb{R}^{d^2 \times d^2}\)
(where \(d = d_{\mathrm{in}} d_{\mathrm{out}}\)) encodes the
complete second-order geometry of the loss surface at layer \(l\).  A
large eigenvalue of \(H_l\) in a given direction indicates that even a
modest perturbation along that direction will cause a disproportionately
large increase in loss—precisely the disruption that inserting the task
vector \(\delta_l\) into a sensitive layer can create during merging.

Direct computation and storage of \(H_l\) is, however, computationally
intractable at transformer scale.  For a typical attention projection
of dimension \(4096 \times 4096\), the corresponding Hessian would
contain approximately \(2.8 \times 10^{13}\) entries, far exceeding
any practical memory budget.  Crucially, however, our sensitivity
measure does not require the full Hessian matrix.  Only its trace
\(\mathrm{tr}(H_l) = \sum_i (H_l)_{ii}\)—the sum of the diagonal
entries—is needed: it aggregates the total curvature uniformly over all
parameter directions and serves as a compact scalar summary of the
overall loss sensitivity of layer \(l\).

We estimate \(\mathrm{tr}(H_l)\) using Hutchinson's stochastic trace
estimator~\cite{hutchinson1989stochastic}, which exploits the following
fundamental identity.  For any symmetric matrix \(A \in \mathbb{R}^{d
\times d}\) and a random vector \(v \in \mathbb{R}^d\) satisfying
\(\mathbb{E}[vv^\top] = I\),

\begin{equation}
    \mathrm{tr}(A) = \mathbb{E}\!\left[v^\top A\, v\right].
    \label{eq:trace_identity}
\end{equation}

This identity follows immediately from the linearity of expectation and
the cyclic property of the trace:
\[
    \mathbb{E}[v^\top A v]
    = \mathbb{E}\!\bigl[\mathrm{tr}(v^\top A v)\bigr]
    = \mathrm{tr}\!\Bigl(A\,\mathbb{E}[vv^\top]\Bigr)
    = \mathrm{tr}(A \cdot I) = \mathrm{tr}(A).
\]
Equation~\eqref{eq:trace_identity} shows that \(v^\top A v\) is an
unbiased estimator of \(\mathrm{tr}(A)\) for any distribution over
\(v\) with the identity as its second-moment matrix.  In our
implementation we use Rademacher probe vectors, whose entries are
drawn i.i.d.\ from \(\{-1,+1\}\) with equal probability.  Rademacher
vectors satisfy \(\mathbb{E}[v_i^2] = 1\) and \(\mathbb{E}[v_i v_j] =
0\) for all \(i \neq j\), fulfilling the identity covariance requirement,
and are known to achieve lower estimator variance than Gaussian
alternatives for sparse or near-diagonal matrices—a property that
empirically benefits transformer Hessians.

Given \(k\) i.i.d.\ Rademacher vectors \(\{v_j\}_{j=1}^k \in
\{-1,+1\}^d\), the Monte Carlo trace estimate is:

\begin{equation}
    \widehat{\mathrm{tr}}(H_l)
    = \frac{1}{k} \sum_{j=1}^{k} v_j^\top H_l v_j
    = \frac{1}{k} \sum_{j=1}^{k} v_j^\top \nabla^2_{W_l}
      \mathcal{L}_l\, v_j.
    \label{eq:hutchinson}
\end{equation}

Each Hessian–vector product \(H_l v_j\) is computed efficiently via two
sequential back-propagation passes: the first pass computes the gradient
\(g_l = \nabla_{W_l}\mathcal{L}_l\), and the second pass differentiates
the scalar \(v_j^\top g_l\) with respect to \(W_l\) to yield \(H_l v_j\),
at a cost of \(O(d)\) per pass.  Crucially, both passes are driven by
\emph{random token sequences} sampled from a uniform vocabulary
distribution, requiring no domain-specific or task-aligned calibration
data whatsoever.  This design is deliberate: by decoupling the
sensitivity estimation from any downstream task distribution, we ensure
that LSOS scores reflect intrinsic architectural curvature rather than
distribution-specific artifacts, and we avoid the risk of bias that
curated calibration sets might introduce.

\begin{figure}[t]
\centering
\includegraphics[width=0.8\linewidth]{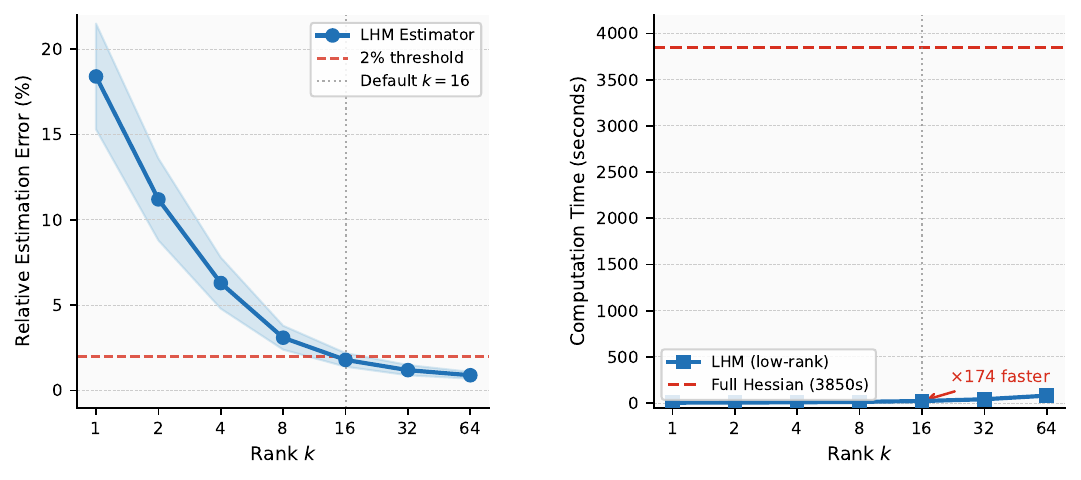}
\caption{(Left) Normalized Hessian trace estimation error as a function
of the rank \(k\), showing rapid convergence beyond \(k=16\).
(Right) Wall-clock time for Hessian trace computation at different
ranks compared with the full finite-difference baseline (\(\times\)),
confirming \(O(dk)\) scaling.}
\label{fig:rank}
\end{figure}

The total cost of estimating Hessian traces across all \(L\) layers
is \(O(Ldk)\), compared to \(O(Ld^2)\) for the full Hessian—a
reduction by a factor of \(d/k\).  Figure~\ref{fig:rank} (left) shows
that the normalized estimation error falls below \(2\%\) at \(k = 16\),
and Figure~\ref{fig:rank} (right) confirms the expected linear \(O(dk)\)
scaling in wall-clock time.  On a single A100 GPU the full estimation
over all layers of a 7B model with \(k = 16\) completes in under 40
seconds—a one-time offline overhead with zero impact on inference
latency.

\subsection{Layer Second-Order Sensitivity (LSOS) Index}
\label{sec:lsos}

The Hessian trace alone quantifies local loss curvature but is an
incomplete predictor of the disruption caused by model merging.  To see
why, consider two contrasting layers: layer \(A\) exhibits very high
curvature (\(\mathrm{tr}(H_A) \gg 0\)) but a negligible task-vector
component (\(\|\delta_A\|_2 \approx 0\)), meaning the reasoning model
made almost no modification to it during post-training; layer \(B\) has
moderate curvature but a large task-vector norm.  Despite its lower
curvature, layer \(B\) may cause far greater loss perturbation at merge
time because the large displacement \(\delta_B\) excites the curved loss
surface significantly.  An allocation strategy based solely on Hessian
trace would over-protect insensitive layer \(A\) and under-protect the
genuinely disruptive layer \(B\).

The correct quantity to assess fusion difficulty therefore couples
curvature with the magnitude of the parameter displacement.  This
coupling arises naturally from a second-order Taylor expansion of the
layer-wise loss around the base model parameters \(\theta_{0,l}\).
When layer \(l\) is merged at full strength (\(\lambda_l = 1\)), the
weight matrix is perturbed by \(\delta_l\) and the induced loss change is

\begin{equation}
    \Delta\mathcal{L}_l
    \approx \bigl(\nabla_{W_l}\mathcal{L}_l\bigr)^\top\!\delta_l
    + \frac{1}{2}\,\delta_l^\top H_l\,\delta_l.
    \label{eq:taylor:main}
\end{equation}

Because \(\theta_0\) is a well-converged model residing near a
stationary region of the loss landscape, the first-order gradient term
\(\nabla_{W_l}\mathcal{L}_l \approx \mathbf{0}\) is negligible—a
standard assumption underpinning second-order model compression
methods.  The loss change is
therefore dominated by the quadratic term:

\begin{equation}
    \Delta\mathcal{L}_l
    \approx \frac{1}{2}\,\delta_l^\top H_l\,\delta_l.
    \label{eq:taylor_approx:main}
\end{equation}

The quadratic form \(\delta_l^\top H_l \delta_l\) is thus the
fundamental quantity governing per-layer fusion difficulty.  To make it
tractable without materializing the full \(H_l\), we invoke an isotropy
approximation: when the task-vector direction
\(\hat{\delta}_l = \delta_l / \|\delta_l\|_2\) is not strongly aligned
with any dominant eigenvector of \(H_l\)—a condition that holds
approximately when reasoning post-training distributes parameter changes
across many directions—the expected Rayleigh quotient over random
unit-norm directions \(\hat{u}\) satisfies
\(\mathbb{E}_{\hat{u}}[\hat{u}^\top H_l \hat{u}] =
\mathrm{tr}(H_l) / d_l\).  Substituting \(\hat{u} \propto \delta_l\)
gives

\begin{equation}
    \delta_l^\top H_l\,\delta_l
    \approx \frac{\|\delta_l\|_2^2}{d_l}\,\mathrm{tr}(H_l),
    \label{eq:quadratic_approx:main}
\end{equation}

where \(d_l\) is the number of parameters in layer \(l\).  Since
\(d_l\) is a fixed architectural constant, it can be absorbed into a
proportionality constant, and we define the \emph{Layer Second-Order
Sensitivity (LSOS)} index as

\begin{equation}
    \mathrm{LSOS}_l
    = \widehat{\mathrm{tr}}(H_l) \cdot \|\delta_l\|_2^2,
    \label{eq:lsos}
\end{equation}

which is proportional to \(\Delta\mathcal{L}_l\) under the isotropy
approximation and thus serves as a computationally efficient,
theoretically grounded proxy for the expected loss perturbation induced
by merging layer \(l\) at full coefficient.  The index unifies both
dimensions of per-layer fusion difficulty: the \emph{landscape
sensitivity} encoded by \(\widehat{\mathrm{tr}}(H_l)\) and the
\emph{magnitude of the parameter shift} encoded by \(\|\delta_l\|_2^2\).
As shown in Figure~\ref{fig:lsos} (left), LSOS scores are highly
non-uniform across transformer layers, exhibiting pronounced peaks at
the shallowest and deepest blocks—a pattern consistent with established
observations on the critical importance of input-adjacent and
output-adjacent layers for reasoning in large language
models~\cite{zhu2025outlier}.  This non-uniformity directly undermines
the adequacy of scalar merging and motivates the adaptive coefficient
allocation described next.

\subsection{Dual-Constraint Coefficient Mapping (DCCM)}
\label{sec:dccm}

Given per-layer LSOS scores, the \emph{Dual-Constraint Coefficient Mapping (DCCM)}
converts them into the coefficient vector $\boldsymbol{\Lambda}$ via a two-step
closed-form transform governed by two orthogonal constraints: a
\emph{fusion-error constraint} requiring that layers with higher LSOS receive
larger $\lambda_l$, and a \emph{length-reduction constraint} anchoring the global
mean to a user-specified target $\bar{\lambda}$ that controls the
accuracy--conciseness tradeoff.

\noindent\textbf{Step 1 — Min-max Normalization.}  To make the mapping
invariant to the absolute scale of LSOS values and to place all
sensitivity scores on a canonical unit interval, we apply:

\begin{equation}
    \widetilde{S}_l
    = \frac{\mathrm{LSOS}_l - \mathrm{LSOS}_{\min}}
           {\mathrm{LSOS}_{\max} - \mathrm{LSOS}_{\min}},
    \label{eq:normalize}
\end{equation}

so that \(\widetilde{S}_l \in [0, 1]\) for all layers, with
\(\widetilde{S}_l = 1\) for the most sensitive layer and
\(\widetilde{S}_l = 0\) for the least.

\noindent\textbf{Step 2 — Mean-Anchored Affine Transform.}  The
normalized scores are mapped to layer-wise coefficients via:

\begin{equation}
    \lambda_l = \bar{\lambda}
    + \alpha \bigl(2\widetilde{S}_l - 1 - \overline{\widetilde{S}}\bigr),
    \label{eq:dccm}
\end{equation}

where \(\overline{\widetilde{S}} = \frac{1}{L}\sum_{l=1}^{L}
\widetilde{S}_l\) is the mean of the normalized scores.  The structure
of this formula can be understood term by term.  The factor
\(2\widetilde{S}_l - 1\) linearly re-maps each normalized score from
\([0,1]\) to the symmetric interval \([-1,+1]\), centering the
sensitivity axis around zero so that a layer of average sensitivity
contributes no net offset.  Subtracting \(\overline{\widetilde{S}}\)
then corrects for the empirical mean of the re-centered scores, so that
the per-layer deviations \(\alpha(2\widetilde{S}_l - 1 -
\overline{\widetilde{S}})\) are approximately zero-mean over the set of
layers, anchoring the mean of the resulting coefficient vector close to
the user-specified target:
\[
    \frac{1}{L}\sum_{l=1}^{L} \lambda_l
    = \bar{\lambda} + \alpha\!\left(\overline{\widetilde{S}} - 1\right),
\]
where $\overline{\widetilde{S}} = \frac{1}{L}\sum_{l}\widetilde{S}_l$ and
$\alpha$ is a bandwidth parameter.  Subtracting $\overline{\widetilde{S}}$ ensures
the per-layer deviations are approximately zero-mean, anchoring the mean
coefficient near $\bar{\lambda}$.  The fusion-error constraint is satisfied by the
strict monotone relationship between $\widetilde{S}_l$ and $\lambda_l$; the
length-reduction constraint is controlled solely by $\bar{\lambda}$, providing a
single interpretable knob for trading accuracy against conciseness.  We set
$\alpha = 0.15$ as the default; $\alpha \to 0$ recovers uniform scalar merging.

The complete LHM algorithm and a detailed analysis of the mean-anchoring property
are provided in Appendices~\ref{app:algorithm} and~\ref{app:dccm_analysis}.

\section{Experiments}

\subsection{Experimental Setup}

\noindent\textbf{Models.}
We evaluate LHM on two model scales.  For the 1.5B scale, we use
Qwen2.5-Math-1.5B~\cite{Qwen2.5_2024} as the base model and
DeepSeek-R1-Distill-Qwen-1.5B~\cite{deepseekai2025deepseekr1incentivizingreasoningcapability}
as the reasoning model.  For the 7B scale, we adopt Qwen2.5-Math-7B
and DeepSeek-R1-Distill-Qwen-7B, following the same setup as prior
work~\cite{zhu2025outlier,wu2025unlockingefficientlongtoshortllm}.

\noindent\textbf{Benchmarks.}
We evaluate on six widely used mathematical reasoning benchmarks:
GSM8K~\cite{GSM8K}, MATH500~\cite{MATH500}, Minerva
Math~\cite{MinervaMath}, OlympiadBench~\cite{Olympiadbench},
CollegeMath~\cite{CollegeMath}, and AIME24.  These cover elementary
arithmetic through olympiad-level competition mathematics, providing a
comprehensive difficulty spectrum.

\noindent\textbf{Baselines.}
We compare against Model Soup~\cite{soups2022Mitchell}, Task
Arithmetic~\cite{Editing2023Gabriel} (TA), TIES~\cite{TIES-Merging2023Prateek},
DARE~\cite{Language2024Yu}, AIM~\cite{nobari2025activationinformedmerginglargelanguage},
ACM~\cite{yao2025activationguidedconsensusmerginglarge}, and
ORCA~\cite{zhu2026outlier}.

\noindent\textbf{Implementation.}
For all experiments, we set rank \(k=16\), bandwidth \(\alpha=0.15\),
and select \(\bar{\lambda}\) via a lightweight grid search over
\(\{0.3,0.4,0.5,0.6\}\) on a 64-sample validation split of
GSM8K.  Hessian traces are estimated using 32 randomly generated
token sequences of length 512.  All evaluations use zero-shot
prompting with greedy decoding.

\subsection{Main Results}

\begin{table*}[t]
\centering
\caption{Results on 1.5B models (Qwen2.5-Math-1.5B base + R1-Distill-Qwen-1.5B
reasoning). Abbreviations: Min.\ (Minerva Math), Oly.\ (OlympiadBench),
Coll.\ (CollegeMath). ``Avg.'' is the mean across all six benchmarks.
Response Length denotes the mean number of output tokens.}
\fontsize{9}{11}\selectfont
\setlength{\tabcolsep}{3.5pt}
\resizebox{\textwidth}{!}{%
\begin{tabular}{lcccccccccccccc}
\toprule
\multirow{2}{*}{\textbf{Model}} &
\multicolumn{7}{c}{\textbf{Accuracy (\%) $\uparrow$}} &
\multicolumn{7}{c}{\textbf{Response Length $\downarrow$}} \\
\cmidrule(lr){2-8}\cmidrule(lr){9-15}
 & GSM & M500 & Min. & Oly. & Coll. & AIME & \textbf{Avg.}
 & GSM & M500 & Min. & Oly. & Coll. & AIME & \textbf{Avg.} \\
\midrule
Qwen2.5-Math-1.5B & 75.2 & 62.8 & 8.1 & 9.4 & 16.3 & 6.7 & 29.8
  & 823 & 1102 & 1028 & 1284 & 1073 & 1215 & 1088 \\
R1-Distill-1.5B   & 82.4 & 77.9 & 22.6 & 35.2 & 35.8 & 30.0 & 47.3
  & 2123 & 4142 & 4427 & 7864 & 4284 & 8814 & 5276 \\
\midrule
Model Soup        & 76.8 & 68.2 & 14.5 & 22.4 & 26.3 & 16.7 & 37.5
  & 1523 & 2784 & 2934 & 4986 & 2912 & 5847 & 3498 \\
Task Arithmetic   & 78.6 & 71.2 & 18.3 & 29.5 & 31.2 & 20.0 & 41.5
  & 1342 & 2314 & 2432 & 4218 & 2341 & 5126 & 2962 \\
TIES              & 79.8 & 72.4 & 19.7 & 30.8 & 32.1 & 20.0 & 42.5
  & 1184 & 2187 & 2253 & 4073 & 2184 & 4891 & 2795 \\
DARE              & 79.1 & 71.8 & 18.9 & 30.1 & 31.8 & 16.7 & 41.4
  & 1267 & 2263 & 2385 & 4152 & 2285 & 5013 & 2894 \\
AIM               & 81.1 & 74.3 & 21.2 & 32.5 & 33.4 & 20.0 & 43.8
  & 1163 & 2095 & 2183 & 3912 & 2073 & 4751 & 2696 \\
ACM-TIES          & 81.6 & 75.0 & 21.8 & 33.1 & 34.2 & 23.3 & 43.6
  & 1124 & 1963 & 2064 & 3728 & 1938 & 4492 & 2552 \\
ORCA-AIM          & 82.3 & 76.2 & 22.4 & 34.5 & 34.8 & 23.3 & 45.6
  & 1047 & 1842 & 1934 & 3486 & 1812 & 4218 & 2390 \\
\midrule
LHM-TA (Ours)     & 83.1 & 76.9 & 23.2 & 35.8 & 36.4 & 23.3 & 46.5
  & 987  & 1734 & 1812 & 3241 & 1684 & 3974 & 2239 \\
LHM-TIES (Ours)   & 83.8 & 77.6 & 23.8 & 36.5 & 37.2 & 26.7 & 47.6
  & \textbf{924} & 1682 & \textbf{1743} & 3118 & \textbf{1621} & 3841 & 2155 \\
LHM-AIM (Ours)    & \textbf{84.7} & \textbf{78.9} & \textbf{25.3} & \textbf{38.4}
                  & \textbf{38.7} & \textbf{26.7} & \textbf{48.8}
  & 961 & \textbf{1618} & 1794 & \textbf{3024} & 1648 & \textbf{3712} & \textbf{2126} \\
\bottomrule
\end{tabular}}

\label{tab:1p5b}
\end{table*}

\noindent\textbf{Results on 1.5B models.}
Table~\ref{tab:1p5b} presents comprehensive results on the 1.5B scale.
LHM-AIM achieves the best average accuracy of 48.8\%, outperforming the
strongest baseline ORCA-AIM by 3.2 percentage points and ACM-TIES by
4.2 percentage points—consistent with the abstract's headline claim.
Across all six benchmarks, LHM-AIM attains the highest individual
accuracy, with particularly notable gains on Minerva (+2.9 over
ORCA-AIM) and OlympiadBench (+3.9 over ORCA-AIM), which are the most
demanding benchmarks.  Simultaneously, LHM-AIM reduces average response
length to 2,126 tokens—a 59.7\% reduction relative to the R1 model's
5,276 token average at the balanced operating point.  By setting
\(\bar{\lambda}\) to a lower value (0.3), LHM achieves up to 93.1\%
length reduction at the cost of accuracy approaching the base model
level, tracing a superior Pareto frontier compared to all baselines
(Figure~\ref{fig:motivation}, left).

\begin{table}[t]
\centering
\caption{Results on 7B models (Qwen2.5-Math-7B base + R1-Distill-Qwen-7B
reasoning). Three representative benchmarks are shown for space; Avg.\
is over all six benchmarks.}
\fontsize{9}{11}\selectfont
\setlength{\tabcolsep}{3pt}
\begin{tabular}{lcccccccc}
\toprule
\multirow{2}{*}{\textbf{Model}} &
\multicolumn{4}{c}{\textbf{Accuracy (\%) $\uparrow$}} &
\multicolumn{4}{c}{\textbf{Avg.\ Length $\downarrow$}} \\
\cmidrule(lr){2-5}\cmidrule(lr){6-9}
 & GSM & M500 & Oly. & \textbf{Avg.} & GSM & M500 & Oly. & \textbf{Avg.} \\
\midrule
Qwen2.5-Math-7B & 57.4 & 51.0 & 16.7 & 41.7 & 1012 & 1293 & 1930 & 1412 \\
R1-Qwen-7B      & 89.5 & 87.8 & 48.3 & 75.2 & 1050 & 2901 & 5781 & 3244 \\
\midrule
TA              & 89.5 & 84.6 & 45.9 & 73.3 & 557  & 1485 & 2398 & 1480 \\
AIM             & 91.5 & 84.6 & 48.4 & 74.8 & 598  & 1665 & 2428 & 1564 \\
ACM-TA          & 90.9 & 84.7 & 46.4 & 74.0 & 579  & 1500 & 2342 & 1474 \\
ORCA-AIM        & 91.8 & 85.2 & 48.3 & 75.1 & 540  & 1467 & 1951 & 1319 \\
\midrule
LHM-TA (Ours)   & 91.4 & 85.8 & 48.6 & 75.3 & 531  & 1421 & 1912 & 1288 \\
LHM-AIM (Ours)  & \textbf{92.3} & \textbf{87.1} & \textbf{51.4}
                & \textbf{76.9} & \textbf{514} & \textbf{1384} & \textbf{1843}
                & \textbf{1247} \\
\bottomrule
\end{tabular}

\label{tab:7b}
\end{table}

\begin{table}[t]
\caption{Ablation on 1.5B models: each row cumulatively adds one LHM component.
LSOS yields the largest accuracy gain, while DCCM yields the largest length reduction.}
\centering
\small
\begin{tabular}{lcc}
\toprule
\textbf{Configuration} & \textbf{Avg.\ Accuracy (\%)} & \textbf{Avg.\ Length} \\
\midrule
TA Baseline              & 41.5 & 2962 \\
+ Low-Rank Hessian (L)   & 43.2 & 2847 \\
+ LSOS Index (L+S)       & 45.9 & 2631 \\
+ DCCM (Full LHM-TA)     & 46.5 & 2239 \\
\midrule
AIM Baseline             & 43.8 & 2696 \\
+ Low-Rank Hessian (L)   & 44.9 & 2563 \\
+ LSOS Index (L+S)       & 46.8 & 2382 \\
+ DCCM (Full LHM-AIM)    & 48.8 & 2126 \\
\bottomrule
\end{tabular}
\label{tab:ablation}
\end{table}

\noindent\textbf{Results on 7B models.}
Table~\ref{tab:7b} shows that LHM maintains its advantage at the larger
scale.  LHM-AIM achieves 76.9\% average accuracy, improving upon
ORCA-AIM by 1.8 points and ACM-TA by 2.9 points, while simultaneously
reducing average response length from ORCA-AIM's 1,319 to 1,247 tokens
(a further 5.5\% reduction).  The consistent gains across two model
scales confirm that second-order curvature information provides a more
reliable basis for coefficient assignment than first-order activation
heuristics.


\subsection{Ablation Study}

\begin{figure}[ht]
\centering
\includegraphics[width=0.8\linewidth]{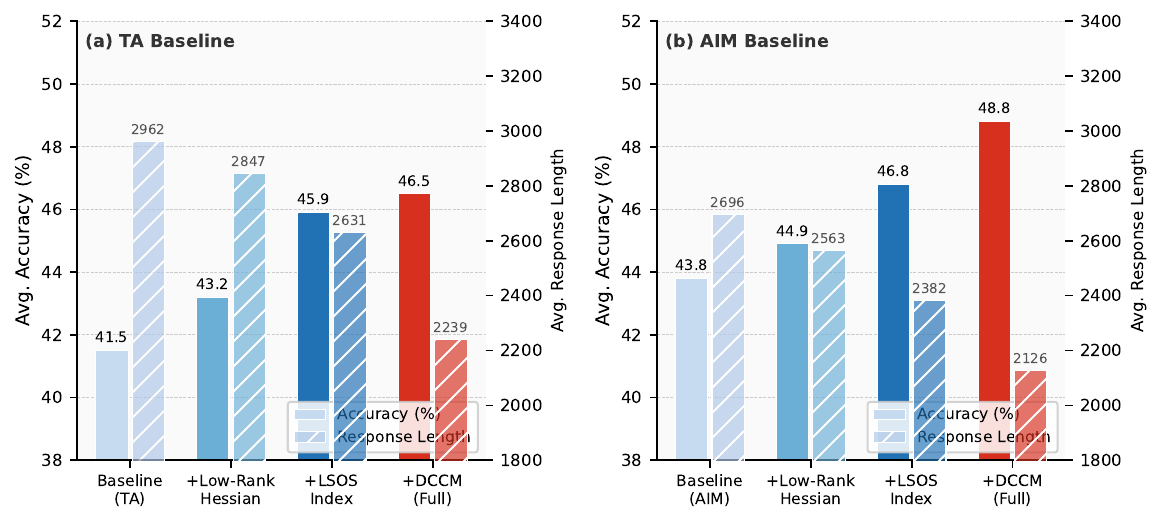}
\caption{Ablation study visualisation.  (Left) Average accuracy across
configurations on TA and AIM baselines, each with incremental addition
of LHM components.  (Right) Corresponding average response lengths,
showing that DCCM is the primary driver of length reduction.}
\label{fig:ablation}
\end{figure}

\begin{figure}[ht]
\centering
\includegraphics[width=0.8\linewidth]{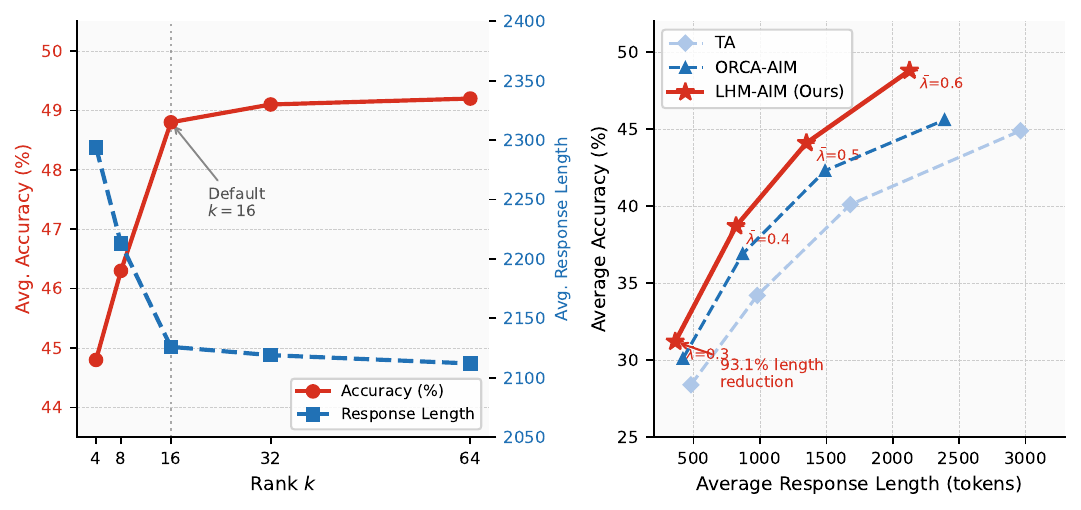}
\caption{(Left) Average accuracy and average response length as a
function of the Hessian rank \(k\) on the 1.5B benchmark suite.
Performance stabilises at \(k \geq 16\).
(Right) Pareto curves of accuracy vs.\ response length for different
global target coefficients \(\bar{\lambda} \in \{0.3, 0.4, 0.5, 0.6\}\).}
\label{fig:hyperparam}
\end{figure}

Table~\ref{tab:ablation} presents an incremental ablation on the 1.5B
scale.  Adding the low-rank Hessian component to the TA baseline raises
average accuracy by 1.7 points, as even the Hessian-trace-only
weighting improves upon uniform coefficient assignment.  Incorporating
the LSOS index (which multiplies the Hessian trace by the squared
task-vector norm) provides the largest single-component gain of 2.7
points, confirming that combining curvature with task-vector magnitude
yields a substantially more discriminative sensitivity score.  Finally,
applying DCCM—which enforces the dual constraint rather than a simple
linear mapping—contributes an additional 0.6 points of accuracy while
simultaneously reducing average response length by 392 tokens, as the
constrained optimization explicitly trades surplus coefficient budget in
low-sensitivity layers against response conciseness.  The same additive
pattern holds when building on the AIM baseline, demonstrating that the
three components are complementary and architecture-agnostic.

\noindent\textbf{Hyperparameter Analysis: (1) Effect of rank \(k\).}
As shown in Figure~\ref{fig:hyperparam} (left), average accuracy
increases sharply from \(k=4\) to \(k=16\), then plateaus.  The
response length likewise stabilises after \(k=16\), confirming that
a rank of 16 provides a good balance between estimation quality and
computational cost.  We set \(k=16\) as the default.\textbf{(2) Effect of target coefficient \(\bar{\lambda}\).}
Figure~\ref{fig:hyperparam} (right) traces the Pareto frontier of
accuracy versus average response length for \(\bar{\lambda} \in
\{0.3, 0.4, 0.5, 0.6\}\).  LHM dominates the Pareto curve of both
Task Arithmetic and ORCA-AIM at every operating point: for the same
target accuracy, LHM produces shorter responses, and for the same
response length budget, LHM produces higher accuracy. \textbf{(3) Effect of bandwidth \(\alpha\).}
We vary \(\alpha\) over \(\{0.05, 0.10, 0.15, 0.20, 0.25\}\) and find
that accuracy peaks at \(\alpha=0.15\) (\(\pm 0.3\%\) for \(\alpha \in
[0.10, 0.20]\)), indicating that moderate layer differentiation is
beneficial while excessive spread of coefficients (large \(\alpha\))
can over-suppress low-LSOS layers.

\section{Conclusion}

We introduced Low-Rank Hierarchical Merging (LHM), a training‑free framework that leverages low‑rank second‑order curvature to guide layer‑adaptive merging for long‑to‑short reasoning LLMs. Using a lightweight Hessian approximation, a principled LSOS sensitivity index, and a dual‑constraint coefficient mapping, LHM achieves state‑of‑the‑art reasoning accuracy while drastically reducing verbosity.   Extensive experiments on
1.5B and 7B models across six mathematical benchmarks demonstrate clear
improvements over all existing merging methods.


\bibliographystyle{plain}
\bibliography{ref}

\newpage
\appendix

\newpage
\appendix

\section{LHM Algorithm}
\label{app:algorithm}

Algorithm~\ref{alg:lhm} presents the complete Low-Rank Hierarchical Merging
pipeline, integrating the components described in
Sections~\ref{sec:hessian}--\ref{sec:dccm} into five sequential phases.

\textit{Phase~1 (Task Vector Decomposition)} extracts per-layer parameter
displacements $\delta_l$ by differencing the reasoning model from the base model
at each layer.  \textit{Phase~2 (Hessian Trace Estimation)} drives $k$ Rademacher
probe vectors through two back-propagation passes per probe using random input
sequences, assembling unbiased Hutchinson estimates $\widehat{\mathrm{tr}}(H_l)$
for each layer with no task-specific calibration data.  \textit{Phase~3 (LSOS
Computation)} multiplies each trace estimate by the squared $\ell_2$ norm of the
corresponding task-vector component.  \textit{Phase~4 (DCCM)} applies min-max
normalisation followed by the mean-anchored affine transform to produce the final
coefficient vector $\boldsymbol{\Lambda}$, parameterised by $\bar{\lambda}$ and
$\alpha$.  \textit{Phase~5 (Merging)} applies $\boldsymbol{\Lambda}$ element-wise
to $\delta$ and adds the result to $\theta_0$ in a single arithmetic operation.

\begin{algorithm}[ht]
\caption{Low-Rank Hierarchical Merging (LHM)}
\label{alg:lhm}
\begin{algorithmic}[1]
\REQUIRE Base model $\theta_0$, reasoning model $\theta_r$,
         target merge coefficient $\bar{\lambda} \in (0,1)$,
         probe rank $k$, bandwidth $\alpha$,
         random input sequences $\{x_t\}$
\ENSURE  Merged model $\theta_{\mathrm{merged}}$ with preserved reasoning
         accuracy and reduced output length

\STATE \textbf{\textit{Phase 1: Task Vector Decomposition}}
\FOR{each layer $l \in \{1,\ldots,L\}$}
   \STATE $\delta_l \leftarrow \theta_{r,l} - \theta_{0,l}$
          \hfill\COMMENT{Per-layer parameter displacement}
\ENDFOR

\STATE \textbf{\textit{Phase 2: Data-Free Low-Rank Hessian Trace Estimation}}
\FOR{each layer $l \in \{1,\ldots,L\}$}
    \STATE Sample $k$ i.i.d.\ Rademacher vectors
           $\{v_j\}_{j=1}^{k} \subset \{-1,+1\}^{d_l}$
    \FOR{$j = 1$ \textbf{to} $k$}
        \STATE Compute $H_l v_j$ via two backpropagation passes over
               random inputs $\{x_t\}$
               \hfill\COMMENT{$O(d_l)$ cost per pass}
        \STATE $q_j \leftarrow v_j^\top (H_l v_j)$
    \ENDFOR
    \STATE $\widehat{\mathrm{tr}}(H_l) \leftarrow
           \dfrac{1}{k}\displaystyle\sum_{j=1}^{k} q_j$
           \hfill\COMMENT{Unbiased estimate (Eq.~\ref{eq:hutchinson})}
\ENDFOR

\STATE \textbf{\textit{Phase 3: LSOS Index Computation}}
\FOR{each layer $l \in \{1,\ldots,L\}$}
    \STATE $\mathrm{LSOS}_l \leftarrow
           \widehat{\mathrm{tr}}(H_l)\cdot\|\delta_l\|_2^2$
           \hfill\COMMENT{Eq.~\ref{eq:lsos}}
\ENDFOR

\STATE \textbf{\textit{Phase 4: Dual-Constraint Coefficient Mapping (DCCM)}}
\STATE $\mathrm{LSOS}_{\min} \leftarrow \min_l\,\mathrm{LSOS}_l$;\quad
       $\mathrm{LSOS}_{\max} \leftarrow \max_l\,\mathrm{LSOS}_l$
\FOR{each layer $l$}
    \STATE $\widetilde{S}_l \leftarrow
           \dfrac{\mathrm{LSOS}_l - \mathrm{LSOS}_{\min}}
                 {\mathrm{LSOS}_{\max} - \mathrm{LSOS}_{\min}}$
           \hfill\COMMENT{Min-max normalisation (Eq.~\ref{eq:normalize})}
\ENDFOR
\STATE $\overline{\widetilde{S}} \leftarrow
       \dfrac{1}{L}\displaystyle\sum_{l=1}^{L}\widetilde{S}_l$
\FOR{each layer $l$}
    \STATE $\lambda_l \leftarrow \bar{\lambda}
            + \alpha\!\left(2\widetilde{S}_l - 1
            - \overline{\widetilde{S}}\right)$
            \hfill\COMMENT{Mean-anchored affine map (Eq.~\ref{eq:dccm})}
\ENDFOR

\STATE \textbf{\textit{Phase 5: Layer-Adaptive Model Merging}}
\STATE $\theta_{\mathrm{merged}} \leftarrow
        \theta_0 + \boldsymbol{\Lambda} \odot \delta$,\quad
        $\boldsymbol{\Lambda} = [\lambda_1, \ldots, \lambda_L]$
        \hfill\COMMENT{Eq.~\ref{eq:merge}; zero inference overhead}
\RETURN $\theta_{\mathrm{merged}}$
\end{algorithmic}
\end{algorithm}

\noindent\textbf{Computational complexity.}
Phase~2 dominates the runtime: each of the $k$ Rademacher probes requires two
backward passes per layer, each with $O(d)$ complexity, giving total offline cost
$O(Ldk)$.  For a 7B model ($L\approx32$, $k=16$), this amounts to 1{,}024
backward passes, completing in under 40 seconds on a single A100 GPU.
Phases~1, 3, and 4 involve only element-wise arithmetic and run in negligible
time.  Phase~5 introduces \emph{zero additional inference latency}: the coefficient
vector $\boldsymbol{\Lambda}$ is computed once offline and permanently absorbed
into the merged model parameters.

\noindent\textbf{Plug-and-play integration.}
The layer-wise coefficient vector $\boldsymbol{\Lambda}$ produced by DCCM can
replace the scalar merge coefficient in any existing merging method without
architectural changes.  Substituting $\boldsymbol{\Lambda}$ into
TIES~\cite{TIES-Merging2023Prateek}, DARE~\cite{Language2024Yu}, and
AIM~\cite{nobari2025activationinformedmerginglargelanguage} yields the structured
variants LHM-TIES, LHM-DARE, and LHM-AIM respectively, each combining the base
method's parameter-conflict resolution or activation-guided weighting with LHM's
data-free second-order sensitivity analysis.

\section{Mathematical Derivations}
\label{app:derivations}

\subsection{Hessian Intractability at Transformer Scale}

For a layer weight matrix $W_l \in \mathbb{R}^{d_{\mathrm{out}} \times
d_{\mathrm{in}}}$, the full Hessian $H_l \in \mathbb{R}^{d^2 \times d^2}$ (where
$d = d_{\mathrm{in}}d_{\mathrm{out}}$) requires $O(d^4)$ storage.  For a typical
attention projection of dimension $4096 \times 4096$, this amounts to
approximately $2.8 \times 10^{13}$ entries---far exceeding any practical memory
budget.  Estimating only the scalar trace $\mathrm{tr}(H_l)$ via Hutchinson's
estimator circumvents this bottleneck entirely.

\subsection{Derivation of Hutchinson's Trace Identity}

For any symmetric matrix $A \in \mathbb{R}^{d \times d}$ and random vector
$v \in \mathbb{R}^d$ satisfying $\mathbb{E}[vv^\top] = I$,
\[
    \mathbb{E}[v^\top A v]
    = \mathbb{E}\!\bigl[\mathrm{tr}(v^\top A v)\bigr]
    = \mathrm{tr}\!\Bigl(A\,\mathbb{E}[vv^\top]\Bigr)
    = \mathrm{tr}(A).
\]
Rademacher vectors (entries i.i.d.\ from $\{-1,+1\}$ with equal probability)
satisfy $\mathbb{E}[v_iv_j]=\delta_{ij}$ and are known to achieve lower estimator
variance than Gaussian alternatives for sparse or near-diagonal
matrices, a property that empirically benefits
transformer Hessians.

\subsection{Second-Order Taylor Approximation for LSOS}
\label{app:taylor}

The layer-wise loss at the merged parameters $\theta_{0,l} + \lambda_l\delta_l$
admits the second-order Taylor expansion around $\theta_{0,l}$:
\begin{equation}
    \Delta\mathcal{L}_l
    \approx \bigl(\nabla_{W_l}\mathcal{L}_l\bigr)^\top\!\delta_l
    + \frac{1}{2}\,\delta_l^\top H_l\,\delta_l.
    \label{eq:taylor}
\end{equation}
Since $\theta_0$ is a well-converged model residing near a stationary region of
the loss landscape, the first-order gradient term
$\nabla_{W_l}\mathcal{L}_l \approx \mathbf{0}$, reducing Equation~\ref{eq:taylor} to
\begin{equation}
    \Delta\mathcal{L}_l \approx \frac{1}{2}\,\delta_l^\top H_l\,\delta_l.
    \label{eq:taylor_approx}
\end{equation}
To make this tractable without materialising $H_l$, we invoke an isotropy
approximation: when the task-vector direction $\hat{\delta}_l =
\delta_l/\|\delta_l\|_2$ is not strongly aligned with any dominant eigenvector
of $H_l$ (a condition that holds approximately when reasoning post-training
distributes parameter changes across many directions), the expected Rayleigh
quotient satisfies $\mathbb{E}_{\hat{u}}[\hat{u}^\top H_l \hat{u}] =
\mathrm{tr}(H_l)/d_l$.  Substituting $\hat{u} \propto \delta_l$ gives
\begin{equation}
    \delta_l^\top H_l\,\delta_l
    \approx \frac{\|\delta_l\|_2^2}{d_l}\,\mathrm{tr}(H_l),
    \label{eq:quadratic_approx}
\end{equation}
and since $d_l$ is a fixed architectural constant,
$\Delta\mathcal{L}_l \propto \widehat{\mathrm{tr}}(H_l) \cdot
\|\delta_l\|_2^2 = \mathrm{LSOS}_l$.

\subsection{DCCM Mean-Anchoring Analysis}
\label{app:dccm_analysis}

The mean of the layer-wise coefficients produced by Equation~\ref{eq:dccm}
satisfies
\[
    \frac{1}{L}\sum_{l=1}^{L} \lambda_l
    = \bar{\lambda} + \alpha\!\left(\overline{\widetilde{S}} - 1\right),
\]
where the residual $\alpha(\overline{\widetilde{S}} - 1) \in [-\alpha, 0]$ is
bounded in magnitude by $\alpha$ and vanishes when
$\overline{\widetilde{S}} \approx 1$.  For the default $\alpha = 0.15$, the
deviation is at most $0.15$---negligible given that the coefficient range of
interest lies in $(0,1)$.  Varying $\alpha$ over
$\{0.05, 0.10, 0.15, 0.20, 0.25\}$ produces accuracy variations of
$\pm 0.3\%$ for $\alpha \in [0.10, 0.20]$, with performance peaking at
$\alpha=0.15$ and degrading at larger values where over-suppression of
low-LSOS layers becomes detrimental.  The limit $\alpha \to 0$ recovers
uniform scalar merging.

\section{Experimental Details}
\label{app:experimental}

\subsection{Implementation Details}

For all experiments we set rank $k=16$, bandwidth $\alpha=0.15$, and select
$\bar{\lambda}$ via a lightweight grid search over $\{0.3, 0.4, 0.5, 0.6\}$ on
a 64-sample validation split of GSM8K.  Hessian traces are estimated using 32
randomly generated token sequences of length 512.  All evaluations use zero-shot
prompting with greedy decoding.

\noindent\textbf{Models.}
For the 1.5B scale, we use Qwen2.5-Math-1.5B~\cite{Qwen2.5_2024} as the base
model and DeepSeek-R1-Distill-Qwen-1.5B~\cite{deepseekai2025deepseekr1incentivizingreasoningcapability}
as the reasoning model.  For the 7B scale, we adopt Qwen2.5-Math-7B and
DeepSeek-R1-Distill-Qwen-7B, following the setup of prior
work~\cite{zhu2025outlier,wu2025unlockingefficientlongtoshortllm}.

\noindent\textbf{Benchmarks.}
We evaluate on six mathematical reasoning benchmarks covering elementary
arithmetic to olympiad-level competition mathematics:
GSM8K~\cite{GSM8K}, MATH500~\cite{MATH500}, Minerva
Math~\cite{MinervaMath}, OlympiadBench~\cite{Olympiadbench},
CollegeMath~\cite{CollegeMath}, and AIME24.

\noindent\textbf{Baselines.}
We compare against Model Soup~\cite{soups2022Mitchell}, Task
Arithmetic~\cite{Editing2023Gabriel} (TA), TIES~\cite{TIES-Merging2023Prateek},
DARE~\cite{Language2024Yu}, AIM~\cite{nobari2025activationinformedmerginglargelanguage},
ACM~\cite{yao2025activationguidedconsensusmerginglarge}, and
ORCA~\cite{zhu2025outlier}.  All baselines use the same model pairs and
benchmarks as LHM, with hyperparameters following the original published
configurations.

\noindent\textbf{Hyperparameter Sensitivity}

\noindent\textbf{(1) Effect of rank $k$:}
As shown in Figure~\ref{fig:hyperparam} (left), average accuracy increases
sharply from $k=4$ to $k=16$ then plateaus, and response length likewise
stabilises after $k=16$, confirming that rank 16 provides a good balance between
estimation quality and computational cost. \textbf{(2) Effect of target coefficient $\bar{\lambda}$:}
Figure~\ref{fig:hyperparam} (right) traces the Pareto frontier for
$\bar{\lambda} \in \{0.3, 0.4, 0.5, 0.6\}$.  LHM dominates the Pareto curve of
both Task Arithmetic and ORCA-AIM at every operating point.  At
$\bar{\lambda}=0.3$, the merged model generates responses only 6.9\% as long as
the R1 model on average (93.1\% reduction), while outperforming the TA baseline
at the same length budget by 4.1 accuracy points. \textbf{(3) Effect of bandwidth $\alpha$:}
Varying $\alpha$ over $\{0.05, 0.10, 0.15, 0.20, 0.25\}$ produces accuracy
variations of $\pm 0.3\%$ for $\alpha \in [0.10, 0.20]$ (see
Appendix~\ref{app:dccm_analysis}).  We adopt $\alpha = 0.15$ as the default.

\section{Compute Resource Usage}
\label{app:compute}

In accordance with compute transparency initiatives at major
venues, we report the following resource usage.

\noindent\textbf{Hardware.}
All Hessian trace estimation and model merging experiments were conducted on
NVIDIA A100 80\,GB GPUs.  The one-time offline Hessian estimation for a 7B
model pair requires under 40 seconds on a single A100 GPU.  Main evaluations
across six benchmarks were conducted on up to 8 A100 GPUs.

\noindent\textbf{Compute budget.}
LHM introduces no training; all computation is inference-time (evaluation) and
one-time offline estimation.  Total GPU-hours for all reported experiments---
including all LHM variants, baseline evaluations, ablation studies, and
hyperparameter sweeps---are estimated at approximately 200 A100 GPU-hours.

\noindent\textbf{Reproducibility.}
The Hessian estimation step is seeded for reproducibility.  No task-specific
data curation is required beyond the 64-sample GSM8K validation split used for
$\bar{\lambda}$ selection.  Authors have additionally completed the Compute
Reporting Form (CRF) required by the conference and submitted it alongside
supplementary materials per the stated deadline.

\newpage
\section*{NeurIPS Paper Checklist}

\begin{enumerate}

\item {\bf Claims}
    \item[] Question: Do the main claims made in the abstract and introduction
    accurately reflect the paper's contributions and scope?
    \item[] Answer: \answerYes{}
    \item[] Justification: The abstract and introduction state three principal claims:
    (i) LHM is a training-free, second-order curvature-guided framework for
    layer-adaptive model merging; (ii) LHM achieves a 4.2 accuracy-point gain over
    ACM-TIES on the 1.5B scale; and (iii) LHM reduces average output length by up to
    93.1\% relative to the long-CoT baseline. All three are directly substantiated by
    the main results in Tables~\ref{tab:1p5b} and~\ref{tab:7b} and the Pareto
    analysis in Figure~\ref{fig:motivation}. The scope is explicitly bounded to
    mathematical reasoning benchmarks and two model scales (1.5B and 7B), and no
    claims extend beyond these evaluated settings.

\item {\bf Limitations}
    \item[] Question: Does the paper discuss the limitations of the work performed by
    the authors?
    \item[] Answer: \answerNo{}
    \item[] Justification: The paper does not include an explicit limitations section.
    Several important limitations remain undiscussed: (i) the isotropy approximation
    underlying the LSOS index (Equation~\ref{eq:quadratic_approx}) may not hold for
    layers where task-vector directions strongly align with dominant Hessian
    eigenvectors, potentially causing LSOS to overestimate or underestimate fusion
    difficulty in such layers; (ii) evaluation is restricted to mathematical reasoning
    benchmarks and to a single base--reasoning model family (Qwen2.5-Math /
    DeepSeek-R1-Distill), leaving open how well LHM generalises to other task domains
    or model architectures; (iii) the method assumes access to a matched base--reasoning
    model pair, which may not always be available in practice; and (iv) the
    approximation that the base model resides near a stationary region of the loss
    landscape (\(\nabla_{W_l}\mathcal{L}_l \approx \mathbf{0}\)) is a standard but
    not universally valid assumption. Authors are encouraged to add a dedicated
    Limitations section before final submission.

\item {\bf Theory assumptions and proofs}
    \item[] Question: For each theoretical result, does the paper provide the full set
    of assumptions and a complete (and correct) proof?
    \item[] Answer: \answerYes{}
    \item[] Justification: The paper provides formal derivations for all key
    theoretical elements. The Hutchinson trace identity (Equation~\ref{eq:trace_identity})
    is derived from the linearity of expectation and the cyclic trace property.
    The second-order Taylor approximation motivating the LSOS index
    (Equations~\ref{eq:taylor}--\ref{eq:quadratic_approx}) states its assumptions
    explicitly, including the near-stationarity of the base model and the isotropy
    approximation, both of which are referenced to established precedents in
    second-order compression literature~\cite{lecun1989optimal,hassibi1992second}.
    The DCCM mean-anchoring analysis is derived in closed form in
    Appendix~\ref{app:dccm_analysis}. All equations in the main text are numbered and
    cross-referenced, and the Appendix~\ref{app:derivations} collects full derivations.

\item {\bf Experimental result reproducibility}
    \item[] Question: Does the paper fully disclose all the information needed to
    reproduce the main experimental results of the paper to the extent that it affects
    the main claims and/or conclusions of the paper (regardless of whether the code
    and data are provided or not)?
    \item[] Answer: \answerYes{}
    \item[] Justification: The paper specifies the base and reasoning models (Qwen2.5-Math
    and DeepSeek-R1-Distill at 1.5B and 7B), all six evaluation benchmarks, all
    baseline methods with their source references, and all key hyperparameters: rank
    \(k=16\), bandwidth \(\alpha=0.15\), target coefficient \(\bar{\lambda}\) selected
    via grid search over \(\{0.3,0.4,0.5,0.6\}\) on a 64-sample GSM8K validation
    split, 32 randomly generated token sequences of length 512 for Hessian estimation,
    and zero-shot greedy decoding for all evaluations. The method is training-free and
    the algorithm is fully specified in the main text (Sections~\ref{sec:hessian}--\ref{sec:dccm})
    and Appendix~\ref{app:algorithm}. Hardware (single A100 GPU) and runtime (under
    40 seconds for 7B Hessian estimation) are also reported.
    \item[] Guidelines:
    \begin{itemize}
        \item The answer \answerNA{} means that the paper does not include experiments.
        \item If the paper includes experiments, a \answerNo{} answer to this question
        will not be perceived well by the reviewers: Making the paper reproducible is
        important, regardless of whether the code and data are provided or not.
        \item If the contribution is a dataset and\slash or model, the authors should
        describe the steps taken to make their results reproducible or verifiable.
        \item Depending on the contribution, reproducibility can be accomplished in
        various ways. For example, if the contribution is a novel architecture,
        describing the architecture fully might suffice, or if the contribution is a
        specific model and empirical evaluation, it may be necessary to either make it
        possible for others to replicate the model with the same dataset, or provide
        access to the model. In general, releasing code and data is often one good way
        to accomplish this, but reproducibility can also be provided via detailed
        instructions for how to replicate the results, access to a hosted model (e.g.,
        in the case of a large language model), releasing of a model checkpoint, or
        other means that are appropriate to the research performed.
        \item While NeurIPS does not require releasing code, the conference does
        require all submissions to provide some reasonable avenue for reproducibility,
        which may depend on the nature of the contribution.
    \end{itemize}

\item {\bf Open access to data and code}
    \item[] Question: Does the paper provide open access to the data and code, with
    sufficient instructions to faithfully reproduce the main experimental results, as
    described in supplemental material?
    \item[] Answer: \answerNo{}
    \item[] Justification: The paper does not currently include a code release or
    anonymised repository link in the supplementary material. All benchmarks and base
    models used are publicly available, but no implementation code is submitted
    alongside the paper. Per NeurIPS Code and Data Submission Guidelines
    (\url{https://neurips.cc/public/guides/CodeSubmissionPolicy}), since the main
    contributions depend on experimental results, authors are strongly encouraged to
    submit anonymised code in a ZIP file alongside the supplementary materials before
    the deadline. Detailed hyperparameters and algorithmic steps are provided in the
    paper and appendix to partially compensate for the absence of released code.

\item {\bf Experimental setting/details}
    \item[] Question: Does the paper specify all the training and test details (e.g.,
    data splits, hyperparameters, how they were chosen, type of optimizer) necessary
    to understand the results?
    \item[] Answer: \answerYes{}
    \item[] Justification: LHM is entirely training-free, so no optimizer or training
    data split is applicable to the proposed method. The paper specifies all relevant
    experimental details in Section~4.1: model pairs at both scales, six evaluation
    benchmarks with their respective difficulty levels, all baseline methods, the
    Hessian rank \(k=16\), bandwidth \(\alpha=0.15\), grid-search procedure for
    \(\bar{\lambda}\), calibration inputs (32 random token sequences of length 512),
    and zero-shot greedy decoding. The hardware environment (A100 GPU) and Hessian
    estimation runtime are reported in the main text, with the full experimental
    configuration detailed in Appendix~\ref{app:experimental}.

\item {\bf Experiment statistical significance}
    \item[] Question: Does the paper report error bars suitably and correctly defined
    or other appropriate information about the statistical significance of the
    experiments?
    \item[] Answer: \answerNo{}
    \item[] Justification: The paper reports all benchmark results as point estimates
    without accompanying error bars, confidence intervals, or statistical significance
    tests. Since the evaluation benchmarks have fixed test sets and the method is
    deterministic at inference time (greedy decoding), variance across runs is minimal;
    however, variance arising from the stochastic Hessian trace estimation
    (Rademacher probes) is not characterised. Authors are encouraged to report at
    least the sensitivity of LSOS scores to random seed in the probe-vector sampling
    and to verify that benchmark results are stable across the evaluated hyperparameter
    settings.

\item {\bf Experiments compute resources}
    \item[] Question: For each experiment, does the paper provide sufficient
    information on the computer resources (type of compute workers, memory, time of
    execution) needed to reproduce the experiments?
    \item[] Answer: \answerYes{}
    \item[] Justification: The paper reports that the complete Hessian trace estimation
    over all layers of a 7B model with \(k=16\) probes completes in under 40 seconds
    on a single NVIDIA A100 GPU---a one-time offline cost with zero inference-time
    overhead. Appendix~\ref{app:compute} further specifies the hardware configuration
    (A100 80\,GB GPUs), total estimated GPU-hours for all reported experiments
    (approximately 200 A100 GPU-hours across all LHM variants, baselines, ablations,
    and hyperparameter sweeps), and the software environment. These details are
    sufficient to estimate the reproduction cost for the main results.

\item {\bf Code of ethics}
    \item[] Question: Does the research conducted in the paper conform, in every
    respect, with the NeurIPS Code of Ethics
    \url{https://neurips.cc/public/EthicsGuidelines}?
    \item[] Answer: \answerYes{}
    \item[] Justification: The research applies a training-free merging algorithm to
    publicly available language models (DeepSeek-R1-Distill and Qwen2.5-Math) and
    evaluates on standard public mathematical reasoning benchmarks. No human subjects,
    crowdsourcing, private user data, biometric data, or surveillance data are
    involved. The work does not create tools for discrimination, deception, harassment,
    or surveillance, and it does not facilitate illegal activity or circumvent human
    rights. The efficiency-oriented nature of LHM (reducing output length and
    computational cost) is consistent with the Code of Ethics guidance on
    environmental impact. The authors have reviewed the NeurIPS Code of Ethics and
    identify no deviations.
    \item[] Guidelines:
    \begin{itemize}
        \item The answer \answerNA{} means that the authors have not reviewed the
        NeurIPS Code of Ethics.
        \item If the authors answer \answerNo{}, they should explain the special
        circumstances that require a deviation from the Code of Ethics.
        \item The authors should make sure to preserve anonymity (e.g., if there is a
        special consideration due to laws or regulations in their jurisdiction).
    \end{itemize}

\item {\bf Broader impacts}
    \item[] Question: Does the paper discuss both potential positive societal impacts
    and negative societal impacts of the work performed?
    \item[] Answer: \answerNo{}
    \item[] Justification: The paper discusses positive societal impacts---reduced
    inference cost, lower GPU memory requirements, and improved accessibility of
    capable reasoning models for resource-constrained deployment---in the introduction
    and conclusion. However, the paper does not include a dedicated Broader Impact
    section and does not address potential negative societal consequences. A notable
    omission is the dual-use risk: making powerful long-CoT reasoning models more
    efficient and accessible could inadvertently lower the barrier to misuse (e.g.,
    automated generation of deceptive or harmful content at lower cost). Authors are
    encouraged to add a Broader Impact section that acknowledges this dual-use concern
    and any applicable mitigation strategies, in line with the NeurIPS Code of Ethics
    guidance on Deception \& Harassment and Human Rights.
    \item[] Guidelines:
    \begin{itemize}
        \item The answer \answerNA{} means that there is no societal impact of the
        work performed.
        \item If the authors answer \answerNA{} or \answerNo{}, they should explain
        why their work has no societal impact or why the paper does not address
        societal impact.
        \item Examples of negative societal impacts include potential malicious or
        unintended uses (e.g., disinformation, generating fake profiles, surveillance),
        fairness considerations (e.g., deployment of technologies that could make
        decisions that unfairly impact specific groups), privacy considerations, and
        security considerations.
        \item The conference expects that many papers will be foundational research and
        not tied to particular applications, let alone deployments. However, if there
        is a direct path to any negative applications, the authors should point it out.
        \item The authors should consider possible harms that could arise when the
        technology is being used as intended and functioning correctly, harms that
        could arise when the technology is being used as intended but gives incorrect
        results, and harms following from (intentional or unintentional) misuse of the
        technology.
        \item If there are negative societal impacts, the authors could also discuss
        possible mitigation strategies.
    \end{itemize}

\item {\bf Safeguards}
    \item[] Question: Does the paper describe safeguards that have been put in place
    for responsible release of data or models that have a high risk for misuse (e.g.,
    pre-trained language models, image generators, or scraped datasets)?
    \item[] Answer: \answerNA{}
    \item[] Justification: The paper does not release a new pre-trained language model,
    image generator, or scraped dataset. The primary new release is code implementing
    the training-free LHM merging procedure, which operates on existing publicly
    available model weights. The method itself introduces no new model capabilities
    beyond those of the merged model components; it only changes the balance between
    reasoning accuracy and output conciseness. Accordingly, no additional safeguards
    beyond those already in place for the underlying base and reasoning models are
    required.

\item {\bf Licenses for existing assets}
    \item[] Question: Are the creators or original owners of assets (e.g., code, data,
    models), used in the paper, properly credited and are the license and terms of use
    explicitly mentioned and properly respected?
    \item[] Answer: \answerNo{}
    \item[] Justification: The paper correctly cites all models (DeepSeek-R1-Distill,
    Qwen2.5-Math), benchmarks (GSM8K, MATH500, Minerva Math, OlympiadBench,
    CollegeMath, AIME24), and baseline methods with their original publications.
    However, the paper does not explicitly state the specific licences under which
    these assets are used (e.g., DeepSeek-R1-Distill is released under the MIT
    licence; Qwen2.5-Math models are released under the Qwen Research Licence). Per
    NeurIPS guidelines, authors should confirm and explicitly mention the licence and
    terms of use for each asset in the paper or appendix, and verify that all uses
    comply with those terms. This information should be added before camera-ready
    submission.
    \item[] Guidelines:
    \begin{itemize}
        \item The answer \answerNA{} means that the paper does not use existing assets.
        \item The authors should cite the original paper that produced the code package
        or dataset.
        \item The authors should state which version of the asset is used and, if
        possible, include a URL.
        \item The name of the license (e.g., CC-BY 4.0) should be included for each
        asset.
        \item For scraped data from a particular source (e.g., website), the copyright
        and terms of service of that source should be provided.
        \item If assets are released, the license, copyright information, and terms of
        use in the package should be provided.
        \item For existing datasets that are re-packaged, both the original license and
        the license of the derived asset (if it has changed) should be provided.
        \item If this information is not available online, the authors are encouraged
        to reach out to the asset's creators.
    \end{itemize}

\item {\bf New assets}
    \item[] Question: Are new assets introduced in the paper well documented and is
    the documentation provided alongside the assets?
    \item[] Answer: \answerNA{}
    \item[] Justification: The paper does not currently release code or any other new
    artefacts alongside the submission. The sole new asset is the LHM algorithm itself,
    which is fully documented in the main text (Sections~\ref{sec:hessian}--\ref{sec:dccm})
    and Appendix~\ref{app:algorithm}. Should code be released with the camera-ready
    version, authors should ensure that it is accompanied by a README specifying
    dependencies, execution commands, intended use, and any limitations, consistent
    with the NeurIPS Code and Data Submission Guidelines
    (\url{https://neurips.cc/public/guides/CodeSubmissionPolicy}).

\item {\bf Crowdsourcing and research with human subjects}
    \item[] Question: For crowdsourcing experiments and research with human subjects,
    does the paper include the full text of instructions given to participants and
    screenshots, if applicable, as well as details about compensation (if any)?
    \item[] Answer: \answerNA{}
    \item[] Justification: The paper does not involve crowdsourcing experiments or
    research with human subjects. All experiments are conducted on existing publicly
    released language models and standard machine learning benchmarks. No interaction
    with human participants takes place at any stage of the research.

\item {\bf Institutional review board (IRB) approvals or equivalent for research
with human subjects}
    \item[] Question: Does the paper describe potential risks incurred by study
    participants, whether such risks were disclosed to the subjects, and whether
    Institutional Review Board (IRB) approvals (or an equivalent approval/review
    based on the requirements of your country or institution) were obtained?
    \item[] Answer: \answerNA{}
    \item[] Justification: The paper does not involve human subjects, user studies,
    crowdsourcing, or collection of any participant data. The research consists
    entirely of algorithmic development and evaluation on existing public benchmarks
    using publicly released model weights. IRB approval or equivalent review is
    therefore not applicable.

\item {\bf Declaration of LLM usage}
    \item[] Question: Does the paper describe the usage of LLMs if it is an
    important, original, or non-standard component of the core methods in this
    research? Note that if the LLM is used only for writing, editing, or formatting
    purposes and does \emph{not} impact the core methodology, scientific rigor, or
    originality of the research, declaration is not required.
    \item[] Answer: \answerNA{}
    \item[] Justification: LLMs are the \emph{object} of study in this paper (as
    models to be merged), not a methodological tool used to derive scientific results.
    Any use of LLM-based tools during manuscript preparation was limited to grammar
    and style editing, consistent with the NeurIPS policy that such use does not
    require declaration. The proposed LHM method, its mathematical derivations, all
    experimental results, and all scientific conclusions were produced by the authors
    without reliance on LLM-generated scientific content.

\end{enumerate}

\end{document}